\documentclass[conference]{IEEEtran}
\IEEEoverridecommandlockouts
\usepackage{cite}
\ifCLASSOPTIONcompsoc
 \usepackage[caption=false,font=normalsize,labelfont=sf,textfont=sf]{subfig}
\else
 \usepackage[caption=false,font=footnotesize]{subfig}
\fi
\usepackage{amsmath}%,amssymb,amsfonts}
\usepackage{graphicx}
\usepackage{xcolor}
\def\BibTeX{{\rm B\kern-.05em{\sc i\kern-.025em b}\kern-.08em
    T\kern-.1667em\lower.7ex\hbox{E}\kern-.125emX}}
\usepackage{hyperref}

\hypersetup{
    colorlinks = true,
    linkcolor  = black,
    filecolor = black,      
    urlcolor = black,
    citecolor=black,
} 

\makeatletter
 \let\old@ps@headings\ps@headings
 \let\old@ps@IEEEtitlepagestyle\ps@IEEEtitlepagestyle
 \def\confheader#1{%
 \def\ps@IEEEtitlepagestyle{%
 \old@ps@IEEEtitlepagestyle%
 \def\@oddhead{\strut\hfill#1\hfill\strut}%
 \def\@evenhead{\strut\hfill#1\hfill\strut}%
 }%
 \ps@headings%
 }
 \makeatother
\confheader{%
2020 11$^{th}$ International Conference on Electrical and Computer Engineering (ICECE)
\hspace{3.6 in}}

 \usepackage[pscoord]{eso-pic}
\newcommand{\placetextbox}[3]{
 \setbox0=\hbox{#3}
 \AddToShipoutPictureFG*{ \put(\LenToUnit{#1\paperwidth},\LenToUnit{#2\paperheight}){\vtop{{\null}\makebox[0pt][c]{#3}}}
 }
 }
 \placetextbox{.21}{0.055}{\small{978-0-7381-1102-5/20/\$31.00 ©2020 IEEE}}

\begin{document}

\title{Traffic Surveillance using Vehicle License Plate
Detection and Recognition in Bangladesh\\
}

%\iffalse % commenting authors
\author{
    \IEEEauthorblockN{\textcolor{white}{.}}
     \hfill
    \and
    \IEEEauthorblockN{Md. Saif Hassan Onim$^1$, Muhaiminul Islam Akash$^2$, Mahmudul Haque$^3$, \\Raiyan Ibne Hafiz$^4$
    \\ \vspace*{-4mm}
    Military Institute of Science \& Technology (MIST), Dhaka - 1216, Bangladesh}\\ 
    
    \small{ \texttt{saif@eece.mist.ac.bd$^1$; miakash3646@gmail.com$^2$; mahmud.eece@gmail.com$^3$; raiyan24r@gmail.com$^4$} }
    \hfill
    \and
    \IEEEauthorblockN{\textcolor{white}{.}}
}
%\fi %commented out authors

\maketitle

\begin{abstract}
Computer vision coupled with Deep Learning (DL) techniques bring out a substantial prospect in the field of traffic control, monitoring and law enforcing activities. This paper presents a YOLOv4 object detection model in which the Convolutional Neural Network (CNN) is trained and tuned for detecting the license plate of the vehicles of Bangladesh and recognizing characters using tesseract from the detected license plates. Here we also present a Graphical User Interface (GUI) based on Tkinter, a python package. The license plate detection model is trained with mean average precision (mAP) of 90.50\% and performed in a single TESLA T4 GPU with an average of 14 frames per second (fps) on real time video footage. 
\end{abstract}


\section{Introduction}
Vehicle surveillance from real time video footage has become an important topic of research in the field of Artificial Intelligence (AI). It is difficult to detect whether all the vehicles are obeying traffic laws, giving tolls by traditional human traffic monitoring and police patrolling. Vehicle License Plate (VLP) detection, recognition and tracking aid forensic experts and practitioners in understanding their interested events. Due to preceding practical applications,  Vehicle detection along with Automatic License Plate Recognition (ALPR) and Optical character recognition (OCR) are now a great interest for researchers \cite{c1}, \cite{c2}. ALPR typically consists of three consecutive steps: detection of VLP, character segmentation and character recognition. A large number of research work in the computer vision field has carried out significant increase in performance specially for availability of Graphical Processing Unit (GPU) and annotated dataset which are available in ImageNet \cite{c3}, Open Images dataset V6+ and other sources. Main deterrents of VLP detection is the diversity of types and templates of VLP in different countries.

A great advancement in object detection is carried out by YOLO stimulated models \cite{c4}, \cite{c5}. New features: Weighted Residual Connections, Cross State Partial connections, Cross Mini-Batch Normalization, Self-adversarial training, Mish Activation are used in YOLOv4 \cite{c6}.Tesseract 4, a latest released stable open-source OCR engine has Unicode (UTF-8) support and it has the ability to recognize more than 100 languages. Tesseract 4 has a new, neural network-based engine specifically Long short-term memory (LSTM) based neural network.

In this paper our intentions were to detect and recognize the VLP of Bangladesh which are of different types and Bangla characters. In order to reach out our intention, we trained a vehicle classifier and a YOLOv4 VLP detection model.
Tesseract version 4 is used as our OCR engine to recognise characters from the detected VLP contour. Before passing the detected VLP contour to the OCR engine, several preprocessing techniques are used to increase the performance of OCR. Another extension of our work is making the whole system more user friendly and easier by designing a GUI.

\section{Literature Review}
An end to end framework is proposed that can super resolve a sequence of low-quality real-world traffic video frames to detect and make the unreadable VLP, readable \cite{c8}.  In many cases, vehicle detection is performed first then VLP detection process starts in the contour of the detected vehicle to reduce false positive detection.

Rayson et al. \cite{c9} proposed a new robust real-time YOLOv2 based CNN model. Their intention was to detect only one class in case of both vehicle and VLP detection. They first detected the vehicle then its VLP. This technique results in reduced number of filters and higher speed. Both Fast YOLO and YOLOv2 were used in their research work that outperformed two commercial ALPR systems in the dataset of SSIG and their proposed UFPR-ALPR. They consecutively achieved ALPR with all correct and redundancy accuracy of 64.89\% and 78.33\% with 35 fps and time of 28.3022 milliseconds.

In \cite{c10} retrieval of text from images of different size, style, orientation, complex background was presented. Bulan et al. \cite{c11} used artificial license plate image generation and unsupervised domain adaptation. Viterbi algorithm was used to identify code sequence from localized VLP region.

\section{Methodologies}
\begin{figure}[t]
\centerline{\includegraphics[width=1\linewidth, height=6cm]{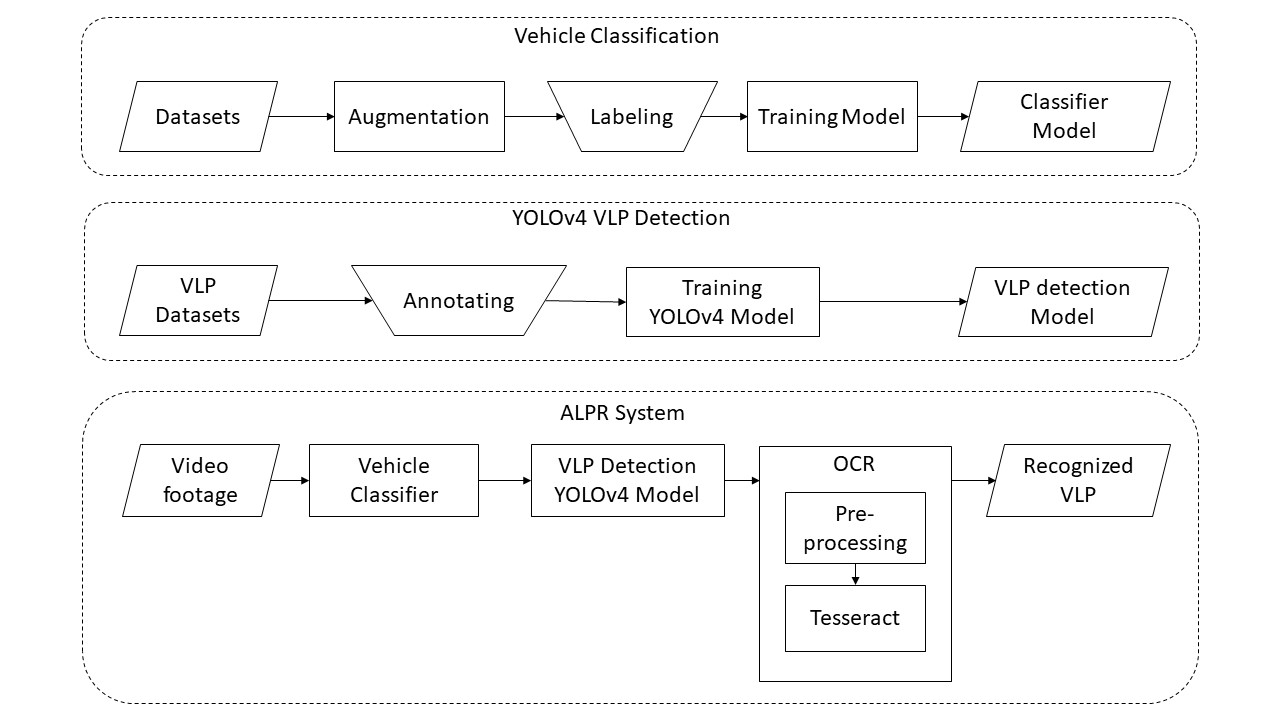}}
\caption{Overall block diagram of the system}
\label{fig1}
\end{figure}

Our ALPR process was comprised of 3 steps in which firstly, we classified video frames with four classes: bus, car, motorbike and truck. If the vehicle classifier detects the vehicle in the video frames, those frames will be fed to the VLP detection model with one class: VLP. This process makes the system efficient. At last OCR system gets the detected VLP contour for character recognition. The complete ALPR system development and deployment is represented in Fig.~\ref{fig1}.

\subsection{Vehicle Classification}
We decided to embed Transfer learning to our CNN architecture as suggested in \cite{c12}. The ResNet based model requires heavy hardware configuration which is not suitable for large scale implementation. Hence, as suggested in \cite{c13} we chose MobileNetV2 as the backbone of our CNN followed by  two custom hidden layers to obtain our desired outcome as it requires relatively lighter hardware configuration with negligible loss in accuracy. For overall implementation of our CNN model we chose TensorFlow (TF) v2.3.0 framework.
\paragraph{Dataset Collection}
We collected images from a video shot in a traffic intersection of Dhaka City, the capital of Bangladesh. The images were collected by cropping the vehicles from video frames. 
\paragraph{Pre-Processing}
We used geometrical data augmentation by using keras to create diversity in the dataset. The collected vehicle images were classified into 'Bus', 'Car', 'Motorbike' and 'Truck'. The collected images had variable sizes. So, they were resized to (96, 96, 3) by up-sampling or down-sampling according to their respective former sizes. All images were chosen after cross-validation and split into training and validation dataset where 6,074 training images being 80\% and 1,519 validation images being 20\% of the total 7,593 images.
\paragraph{Transfer Learning (TL)}
The amount of collected images is not sufficient to get the best outcome from a DNN algorithm. So, to overcome this, along with pre-trained weights from ‘imagenet’ we got additional 2,263,108 parameters. MobileNetV2 was used as the backbone of the DNN architecture which took an input of  shape (96, 96, 3) and gave an output of shape (3, 3, 1280).

\paragraph{Additional Layers}
On top of the MobileNetV2 layers, an average pooling layer was implemented. Finally, we got our desired output using a dense layer of DNN which gives the output i.e. the vehicle classification. The use of these additional layers contributed to 5,124 trainable parameters resulting in a total of 2,257,984 parameters.

\paragraph{Hyper-parameters And Tuning}
The training hyper-parameters are given in TABLE.~\ref{tab1}. Those parameters are chosen after several trails. RMSprop \cite{c14} algorithm was used for optimization and classification in the final stage. Besides, ‘sparse categorical cross entropy’ was used. The last stage categorizes input images into classes they fall under.

\subsection{YOLOv4 VLP Detection}
For the purpose of robust object detection, the neural network’s fast operating speed was the primary aim of YOLOv4. Alexey et al. \cite{c6} in YOLOv4 used CSPDarknet53 \cite{c15} as backbone. It consists of 29 convolutional layers, a $725\times725$ receptive field with parameters of 27.6 million. YOlOv3 \cite{c16} is used as head and SPP \cite{c17}, PAN \cite{c18} played the role of neck of the object detector. Mosaic and self-adversarial training were used in YOLOv4 as new data augmentation techniques.
We trained our custom YOLOv4 \cite{c6} model with an annotated dataset of 1500 training and 300 validation from Open Images Dataset V6+. We obtained substantial results of correctly detecting VLP from real world video footage from the roads in Dhaka, Bangladesh with significant traffic density. If the number of classes = CL and with a view to predicting bounding boxes, anchor boxes is A in YOLO each with four coordinates, confidence and class probabilities \cite{c19}. The number of filters (used in three convolutional layers before YOLO layers) can be represented by equation \eqref{eq1}. Here A = 5 is used. All modified parameters are given in TABLE.~\ref{tab1}.
\begin{equation}
filters = (A+CL)\times3\label{eq1}
\end{equation}

\begin{table}[b]
\caption{Modified Parameters for Vehicle Classification \& YOLOv4 VLP Detection}
    \centering
    \begin{tabular}{l|c|c}
    % \vline
    \hline
    Modified Parameters &	Vehicle Classification &  VLP Detection\\
    \hline
    classes & 4&	1\\
    batch & 32 &	64\\
    subdivision& - &	16\\
    learning\_rate &0.0001  &	0.001\\
    max\_batches (refers &&\\
    
    to epochs) & 100& 	3500\\
    steps & 190, 90&	4800, 5400\\
    width, height & 96, 96 & 416, 416\\
    filters & 32 &	$(5+1)\times3$ = 18\\ 
    \hline
    % \vline
    \end{tabular}
    % \caption{Caption}
    \label{tab1}
\end{table}

Training was started with an initially set iteration number of 6000. After 2000 iterations, the decrease of average loss was not significant, moreover the accuracy line started showing ups and downs after 3000 iterations.The training was capable of taking backup of weights after each 1000 iterations. The performance of every backup weights are given in TABLE.~\ref{tab2}. We took the weight after 3000 iteration as our best weights for testing. The model was trained in Google Colab using a single TESLA T4 GPU with 12 GB RAM.
\iffalse
\begin{figure}[b]
\centerline{\includegraphics[width=0.7\linewidth, height=4cm]{YOLO.png}}
\caption{Training average loss vs iteration and mAP vs iteration.}
\label{fig2}
\end{figure}
\fi
\subsection{VLP Processing And Character Recognition}
OCR is used to recognize characters from raw images in a machine-readable format. There are mainly two steps for text recognition by means of OCR: image pre-processing and OCR itself. To increase the accuracy and efficiency of OCR, the identified license plates that we got from the VLP detection algorithm were binarized using Otsu’s Binarization Method \cite{c20} in order to get desired results despite factors like lighting, background and environment. The processed image is then fed to Tesseract 4. The binary image is initially segmented character-wise by tesseract, and individual characters are recognized. The recognition is mainly focused on Bangla character recognition. Tesseract has provision of using custom trained data for OCR but since the license plates have printed texts in them, we have used the default Bangla trained data which can be obtained through custom installation of Tesseract.

\subsection{GUI}
\begin{figure}[b]
\centerline{\includegraphics[width=0.9\linewidth, height=4cm]{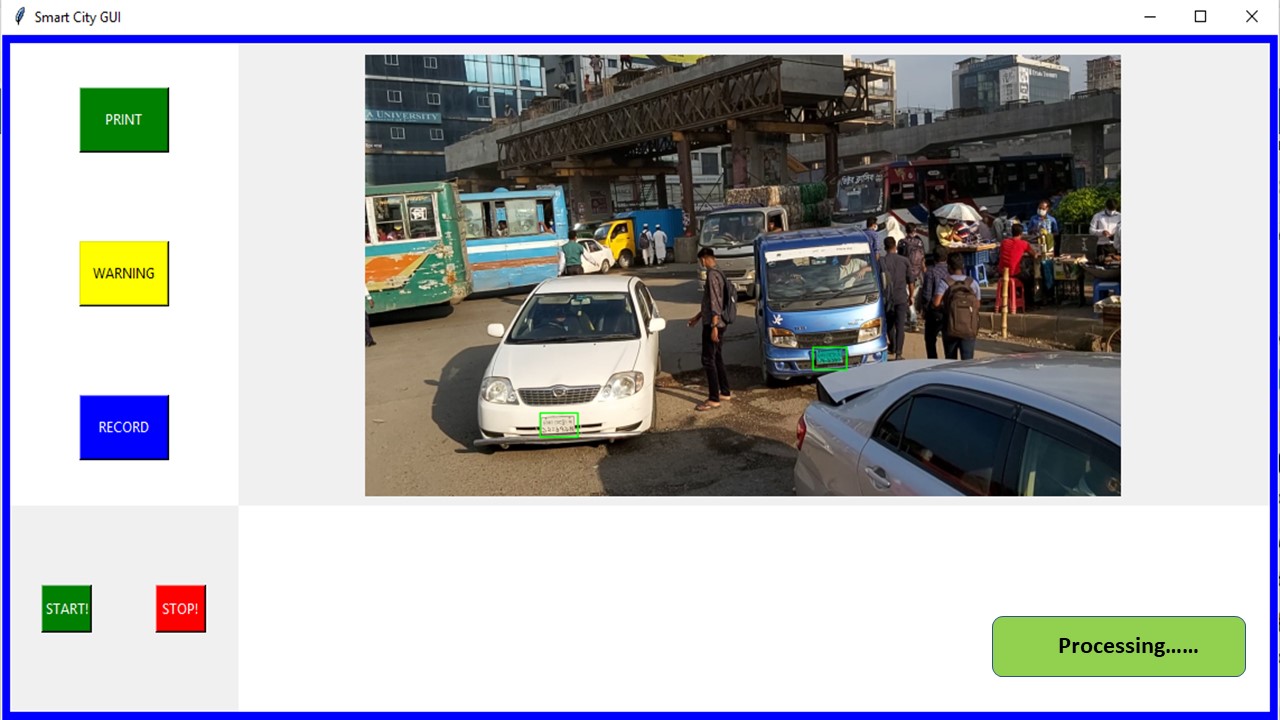}}
\caption{Interface of The GUI}
\label{fig3}
\end{figure}

Our GUI was built using 5 types of widgets. A user interacting with our system using the GUI can perform various functions like monitoring vehicles, recognizing VLP and query details about a vehicle from its license plate number. These functions have been implemented by using various widgets and geometry managers namely label, button, text, pack, frame and grid. In GUI, we used a total of 5 frames. A main frame which encloses 4 other frames. In Fig.~\ref{fig3}, there are a total of 5 buttons on the GUI for performing different functions on command.

\begin{itemize}
    \item Start Button: starts classifying the video footage.
    \item Stop Button: stops classifying the video footage.
    \item Print Button: displays the detected number plate and other information.
    \item Warning Button: triggers an alarm to specific responders. 
    \item Record Button: creates a recording of the processed video footage and saves it on local drive.
\end{itemize}

\iffalse
\subsubsection{Start Button}
Starts classifying the video footage.  
    %  \item
\subsubsection{Stop Button}
Stops classifying the video footage. 
    %  \item
\subsubsection{Print Button}
Displays the detected number plate and other information. 
    %  \item
\subsubsection{Warning Button}
Triggers an alarm to specific responders. 
    %  \item
\subsubsection{Record Button}
Creates a recording of the processed  video footage and saves it on local drive.

\fi

\section{Experimental Result}
\subsection{Vehicle Classification}
Training accuracy and loss of the classification model are 96.43\% and 0.1065 respectively along with validation accuracy of 96.91\% and loss of 0.1023.

\begin{table}[!b]
\caption{Performance Metric of YOLOv4 VLP Detection}
    \centering
    \begin{tabular}{c|c|c|c|c}
    % \vline
    \hline
    
Epochs&mAP&Precision&recall&F1-score\\
    \hline

1000&
78.20&
0.70&
0.81&
0.75\\

2000&
89.57&
0.91&
0.88&
0.89\\

3000&
90.50&
0.93&
0.86&
0.89\\

4000&
88.99&
0.92&
0.84&
0.88\\

5000&
89.02&
0.93&
0.87&
0.90\\

6000&
89.25&
0.93&
0.86&
0.89\\

    \hline
    % \vline
    \end{tabular}
    % \caption{Caption}
    \label{tab2}
\end{table}

\begin{figure}[!b]
\centerline{\includegraphics[width=0.9\linewidth, height=5cm]{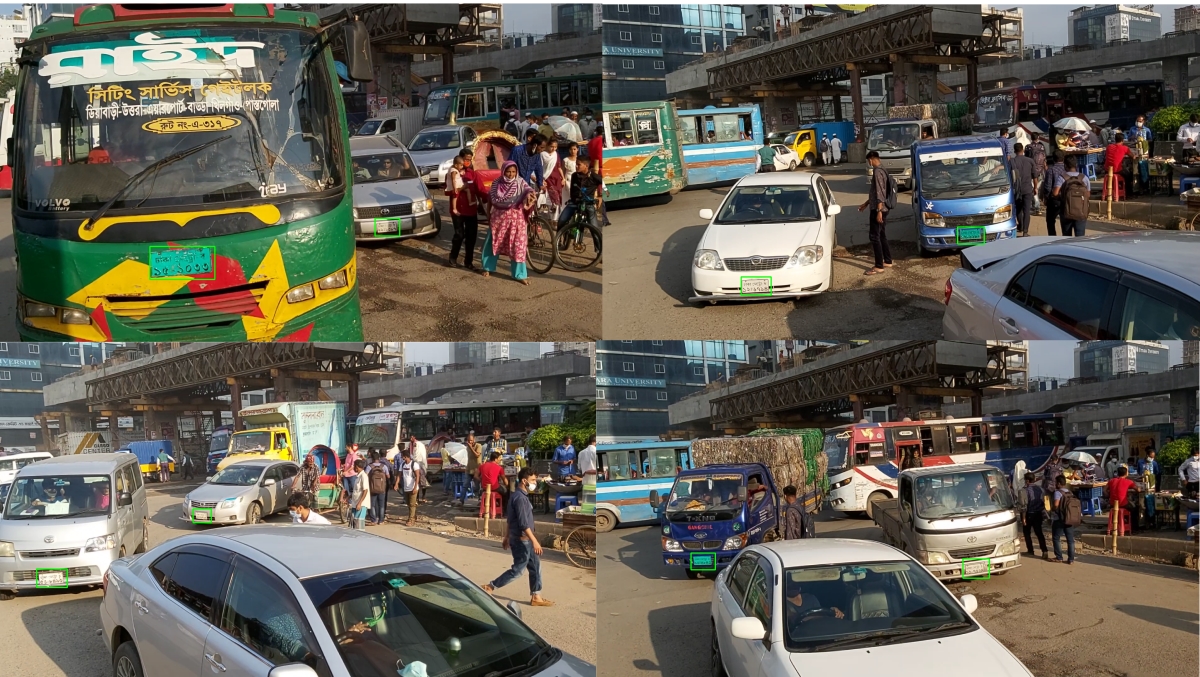}}
\caption{Detected VLP samples from real world captured video footage, Dhaka, Bangladesh.}
\label{fig4}
\end{figure}

\subsection{VLP Detection}
We tested our model on recorded video footage captured by a Samsung Galaxy A70 smartphone with video resolution of $1920\times1080$ with 30 fps. The visual result of VLP detection with drawing bounding boxes are given in Fig.~\ref{fig4}. The attached images represent that our model was capable of detecting multiple VLP in the video frames during daylight. In Google Colab TESLA T4 GPU, our model runs on average 14 fps. The model gained the best mAP of 90.50\% and recall of 0.86 during training and other performance parameters are shown in TABLE.~\ref{tab2}.

\subsection{Character Recognition}
Tesseract's performance was tested with the test dataset available in tesseract forum and the performance we got from the test data set is shown in TABLE.~\ref{tab3}. The histogram of the VLP contour and output of tesseract are illustrated in Fig.~\ref{fig5a} and \ref{fig5b}. Our ALPR system was tested on a recorded real-world traffic video footage from Dhaka, Bangladesh. The VLP detection model detects the VLP contours in the consecutive video frames from the video. Then those detected contours are extracted from the video and passed to the OCR system. Detected VLP contours are of different geometric orientation and brightness. For these reasons, to increase character recognition efficiency, the images are processed before feeding to tesseract.

%  \begin{figure}[t]
% \centerline{\includegraphics[width=0.9\linewidth, height=3cm]{OCR.png}}
% \caption{Pre-processing of detected VLP contour and recognized OCR output}
% \label{fig5}
% \end{figure}
\iffalse
\begin{figure}[t]
\centerline{\includegraphics[width=0.9\linewidth, height=2cm]{FP1.png}}
\caption{False positive detection samples of VLP}
\label{fig6}
\end{figure}
\fi

\begin{table}[!t]
\caption{Performance Analysis of OCR}
    \centering
    \begin{tabular}{c|c|c|c}
    % \vline
    \hline
Image&No of characters&
Accuracy of OCR&
Time taken for OCR \\
 No&
 extracted&(in \%)&(in Seconds)
\\
    \hline
1&5& 56& 0.402\\
2&4& 40& 0.548\\
3&7& 70& 0.402\\
4&4 &40 &0.701\\ 
5&4& 44& 0.705\\ 
6&2 &22 &0.7\\
7&8 &73 &1.717\\
8&6 &67& 0.806\\
9&6& 55 &0.596\\

    \hline
    % \vline
    \end{tabular}
    % \caption{Caption}
    \label{tab3}
\end{table}

\begin{figure}[!t]
\centering
\subfloat[]{\label{fig5a}\includegraphics[width=0.48\linewidth]{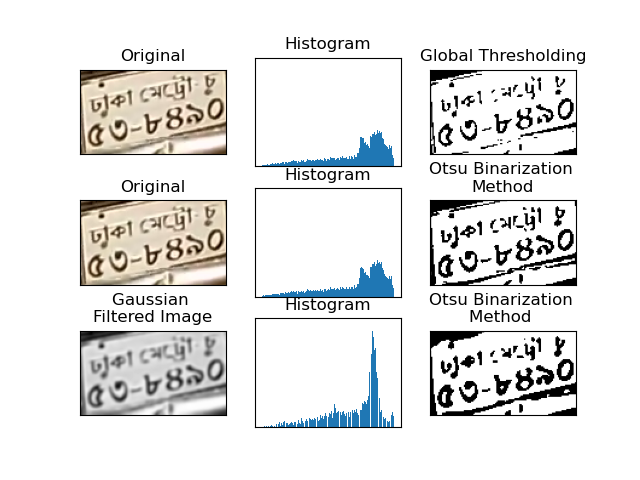}
% \subcaption{}
}\hfil
\subfloat[]{\label{fig5b}\includegraphics[width=0.48\linewidth]{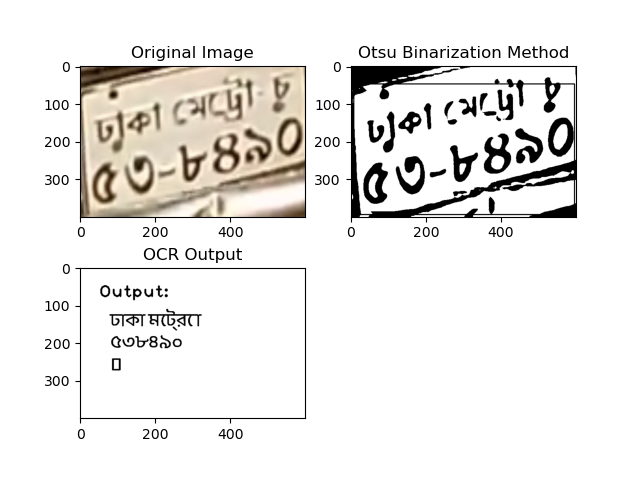}%]
% \label{fig: fig_second_case}
% \subcaption{}

}
\caption{Pre-processing of detected VLP contour and recognized OCR output}
\label{fig5}
\end{figure}

\section{Error Analysis}
 The probability of false positive detection of VLP is not negligible as different objects existing on real world roads might be similar to VLP, this result is false positive detection. If any false positive VLP contour is passed to the OCR system to extract the characters of VLP, the processing will be redundant. Another noticeable failure to detect VLP is when it is under shadow or in direct sunlight. Detecting Vehicle contour first and VLP contour second in the detected vehicle, false positive detection rate can be reduced. Initial representation in Fig.~\ref{fig5b} resembles that Bangla characters output sequence got disrupted. Which was overcome by writing the result of OCR in a text file. The angular orientation of VLP contour is one of the causes of erroneous tesseract output. We noticed that when we manually fixed the orientation, the result got improved. But the main problem of OCR is handling blurry images. Motion blur, distance and angle of camera position with respect to VLP significantly affects the performance of OCR. Performance might be improved if several consecutive VLP frames are processed and filtered to come to a single image with better quality.

 \section{Conclusion}
 This paper emphasized on license plate detection and its character recognition of the vehicles of Bangladesh. A design of GUI is also presented so that it can easily be operable by the users. Our initial works started consecutively from image classification to detect the existence of vehicles in video frames, VLP detection using YOLOv4 and tesseract as OCR engine. Character recognition from VLP was one of the most competitive tasks in this ALPR system due to various orientation, motion blur, lighting condition etc. of detected VLP. As future work, our intentions are to reduce the effects of blurry VLP and to overcome the deterrents of OCR by deploying preprocessing  and compare our prospective model with several existing algorithms from published literature, specially for VLP detection and recognition of vehicles in Bangladesh.

\bibliographystyle{./bibliography/IEEEtran}
\bibliography{./bibliography/IEEEabrv,./bibliography/References}

\end{document}